\documentclass[letterpaper, 10 pt, conference]{ieeeconf}  

\IEEEoverridecommandlockouts                              

\usepackage[utf8]{inputenc}
\usepackage{graphics} 
\usepackage{amsmath} 
\usepackage{amssymb}  
\usepackage{cite}
\usepackage{graphicx} 
\usepackage{hyperref}
\usepackage{epstopdf}
\usepackage{tikz}
\usepackage{tikz-cd}
\usepackage{wasysym}
\usetikzlibrary{shapes,arrows,positioning,calc, fit}
\usetikzlibrary{shapes,arrows,positioning,calc}
\usepackage{grffile} 

\usepackage{amsthm}
\usepackage{bm}
\usepackage{amsbsy}
\usepackage{textcomp}
\usepackage[capitalise]{cleveref}

\usepackage[caption=false,font=footnotesize]{subfig}

\usepackage{etoolbox}
\robustify{\subref}
\title{\LARGE \bf Human-Robot Handovers using Task-Space Quadratic Programming}

\author{
	Mohamed Djeha, 
	Antonin Dallard,
	Ahmed Zermane, 
	Pierre Gergondet \\ and 
	Abderrahmane Kheddar,~\IEEEmembership{Fellow,~IEEE}
	\thanks{The authors are with the CNRS-University of Montpellier LIRMM Interactive Digital Humans group, Montpellier, France. {\tt\small mohamed.djeha@lirmm.fr}}
	\thanks{A. Kheddar and P. Gergondet are also with the CNRS-AIST Joint Robotics Laboratory, IRL, Tsukuba, Japan.}
}

\def\actConf{q}

\def\actJac{J}

\def\setR{{\cal R}}

\def\quatern{\mathbf{q}}

\def\frameWorld{\setR_{\rm w}}
\def\frameGrasp{{\cal R}_{\rm grasp}}
\def\frameEndEff{{\cal R}_{\rm ee}}
\def\frameObs{{\cal R}_{\rm obs}}
\def\frameObj{{\cal R}_{\rm obj}}
\def\trajTrackTaskState{\eta_{\rm tt}}
\def\trajTrackTaskStateDot{\dot{\eta}_{\rm tt}}

\def\observTaskState{\eta_{\rm obs}}
\def\observTaskStateDot{\dot{\eta}_{\rm obs}}

\newcommand{\norm}[1]{\begin{Vmatrix}
		#1
\end{Vmatrix}}
\newcommand{\skewMat}[1]{\begin{bmatrix}
		#1\times
\end{bmatrix}}
\newcommand{\mcrtc}{\texttt{mc\_rtc}}
\newcommand{\inR}{\in \mathbb{R}}

\tikzset{
	block/.style = {draw, fill=white, rectangle, minimum height=1.5em, minimum width=2em},
	tmp/.style  = {coordinate}, 
	sum/.style= {draw, fill=white, circle, node distance=1cm},
	input/.style = {coordinate},
	output/.style= {coordinate},
	pinstyle/.style = {pin edge={to-,thin,black}},
	dotted_block/.style={draw=black!50!white, line width=1.5pt, dash pattern=on 3pt off 3pt on 3pt off 3pt, inner ysep=3mm,inner xsep=2mm, rectangle, rounded corners}
}
\newcounter{mycounter} 	

\makeatletter
\def\endthebibliography{%
	\def\@noitemerr{\@latex@warning{Empty `thebibliography' environment}}%
	\endlist
}
\makeatother
\begin{document}

	\maketitle

	\begin{abstract}
		Bidirectional object handover between a human and a robot enables an important functionality skill in robotic human-centered manufacturing or services. The problem in achieving this skill lies in the capacity of any solution to deal with three important aspects: (i) synchronized timing for the handing over phases; (ii) the handling of object pose constraints; and (iii) understanding the haptic exchanging to seamlessly achieve some steps of the (i). We propose a new approach for (i) and (ii) consisting in explicitly formulating the handover process as constraints in a task-space quadratic programming control framework to achieve implicit time and trajectory encounters. Our method is implemented on Panda robotic arm taking objects from a human operator. 
	\end{abstract}

	\section{Introduction}
	Human-centered robotics trends aim at close contact interactions in different areas: hospitals, industry, malls, homes, restaurants, etc. This cohabitation implies a large taxonomy of human-robot interactions. Among the most challenging one, seamlessly exchanging objects between a human and robot as \emph{giver} and \emph{receiver} agents through (bi-directional) handovers. Indeed, handing objects between humans is a very frequent daily form of interaction in almost all domains~\cite{ortenzi2021tro}. Therefore, embedding human-centered robotic systems in automation and services with handovers capabilities, is a key enabler for rich cognitive interaction~\cite{ajoudani2018autonomousRobots,billard2019science}. Although human-human handover is an intuitive behavior, it is a result of complex and sophisticated learned social-cognitive communication channels~\cite{strabala2013jhrisc}. Grasping such a complexity in an equivalently robot control formulation strategy is not a trivial problem.

	Indeed, how to \emph{codify} the human-robot handover process has been thoroughly addressed in the research community. The state-of-the-art studies split the handover process in two main phases: the \emph{pre-grasping} phase~\cite{shibata1995roman,waldhart2015iros,mainprice2021roman,huber2008roman,shi2013rss,koene2014roman,remazeilles2015rss,bdiwi2013ssd} and the \emph{grasping} phase~\cite{nagata1998icra,chan2013ijrr,medina2016humanoids,solak2019iros,costanzo2021frontiersRob-AI}. Such a breakdown is simply chronological: the former focuses on detecting the object, estimating/predicting the human movement and planning the robot reaching motion toward the meeting point, whereas the latter consists of understanding the interaction forces (haptics) applied at the  object by the two handing over agents during the exchange such that to ensure a stable grasping. In~\cite{medina2016humanoids} a \emph{retraction} phase is considered, which describes how the giver and receiver move away after the latter gets the full control of the object. Yet, for an exhaustive handover survey, please refer to the excellent recent review in~\cite{ortenzi2021tro}.  
	
	Despite the huge amount of works addressing human-robot handover, one issue remains open: how to ensure/codify an adaptive behavior of the robot w.r.t to a human versatile intention on \emph{where} to exchange the object, aka \emph{HandOver Location} (HOL), with the robot, and in which configuration (\emph{orientation}).  
	A large amount of existing works required the knowledge of the HOL as a precondition either for motion planning or control formulations of the pre-grasping phase. In particular, the HOL is often considered either as fixed (the robot moves systematically to a fixed spot to pick-up the object) or pre-planned (real-time detection and estimation of the object fixed-location)~\cite[Tables 1-2]{ortenzi2021tro}. There are works that provide HOL estimation and on-line prediction methods~\cite{prada2014iros,medina2016humanoids,vahrenkamp2016humanoids,maeda2017autonomousRobots,widmann2018ecc,nemlekar2019icra}. However, such existing approaches have at least one of the following shortcomings:
	\begin{itemize}
		\item Collecting data sets to train motion prediction models;
		\item Time-consuming computations leading to a hashed or a slow handover; 
		\item Magic-numbers of meta-parameters to be tuned;
		\item Conservatism in the HOL prediction policy; 
		\item Non-systematic success of the handover performance;
		\item The object orientation at the HOL is often kept the same during the experiments;
		\item Knowledge of the HOL is very often required.
	\end{itemize} 
	\begin{figure}
		\centering
		\includegraphics[width=0.49\columnwidth]{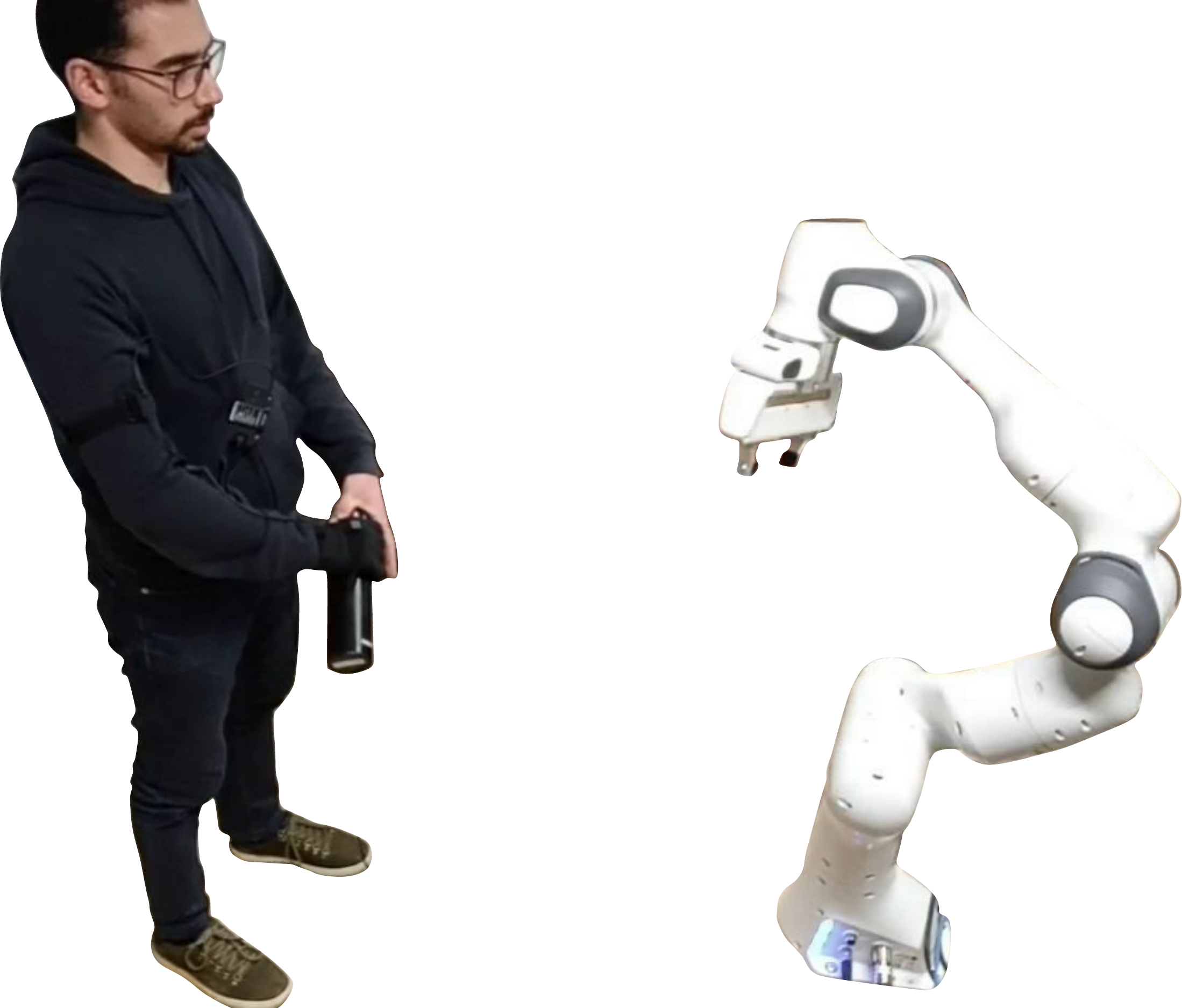}
		\includegraphics[width=0.49\columnwidth]{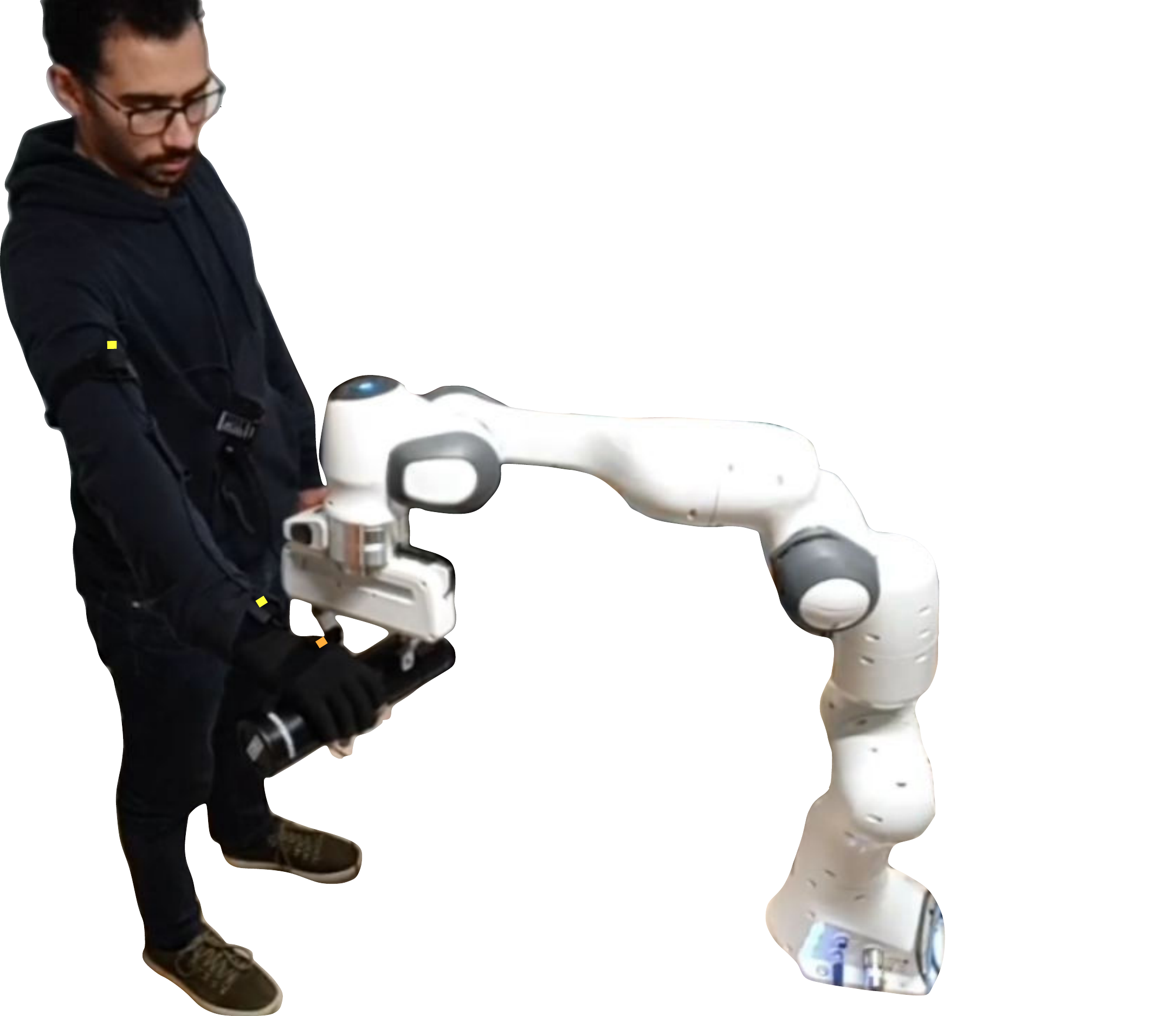}
		\caption{Proposed seamless handover process executed without any prior knowledge about the HOL and final object orientation.}
		\label{fig:proposed handover}
	\end{figure}
	In this work, we formulate the premises of a generic robot control-strategy that ensures high fidelity and reliability of reaching motion for bi-directional handover scenarios and guarantees adaptability w.r.t the HOL and object orientation (\cref{fig:proposed handover}). Particularly, our control formulation gathers the  advantages from both methods in~\cite{costanzo2021frontiersRob-AI,medina2016humanoids}. Indeed, our method is HOL knowledge free. We only assume that the object structural properties and dimensions are known, and there exists a sensor that provides the pose (Cartesian position and orientation) of the object of interest, e.g.~\cite{paolillo2018ral}. Conversely to~\cite{costanzo2021frontiersRob-AI}, our method ensures/codifies a real-time proactive and adaptive robot motion w.r.t to the human versatile intention on \emph{where} to exchange the object with the robot and in which configuration (\emph{orientation}) without the need of offline acquired data.	
	Our aim is to build on our existing multi-objective task-space optimization Quadratic-Programming (QP)  framework~\cite{bouyarmane2018tac,bouyarmane2019tro} that prove to be very powerful in an industrial context~\cite{kheddar2019ram} and a human-assistance context~\cite{bolotnikova2021sii}. In particular, our proposed idea relies on the concept of \emph{coupled tasks} where the state of one task is forwarded as the reference for another task. More precisely, we formulate an \emph{observation task} that estimates either human hand or object full-state in terms of pose, velocity and acceleration which are then forwarded as references for a \emph{trajectory-tracking task} for the robot end-effector to track. Leveraging multi-objective QP  control paradigm, these two tasks run concurrently in a \emph{leader-follower} fashion: the object (leader) moves and converges to the HOL while the robot end-effector (follower) moves proactively toward the object while adjusting its orientation accordingly. From this standpoint, our approach is similar to the coupled Dynamical Systems (DS) in~\cite{medina2016humanoids}. Except that our approach is computationally cheap and does not require the estimation of the HOL nor to separately learn human and robot DS models. 
	To sum-up, our contributions to coupled task-space observer-controller are as follows:
	\begin{itemize}
		
		\item Systematic approach (no conservative policy of HOL prediction) with less parameter to tune (task gains only);
		\item No need for planning and time consuming processes of collecting data and model training; 
		\item Requires no time-consuming computations resulting in a smooth (fluid) and seamless handover motion;
		\item Codifies rich behaviors with multi-objective QP control paradigm: proactive position and orientation control while accounting for safety constraints (collision avoidance, joint and actuation limits, etc.);
		\item Generalizes to bi-directional human-robot handovers;
		\item Needs only a sensor or estimator that provides the object pose, e.g.~\cite{rana2021sj,jongeneel2022hal,paolillo2018ral}. 
	\end{itemize}
	
	\textbf{Notations:} $x^{\star}\in\mathbb{R}^3$ is  expressed in frame $\setR_{\star}$. $R_{\star\star}^{\star}\in SO(3)$ denotes the rotation matrix of the frame $\setR_{\star\star}$ w.r.t $\setR_{\star}$.  If $\setR_{\star}=\frameWorld$, the world-frame,  then $R_{\star\star}^{\star}=R_{\star\star}$. $\skewMat{\omega}={\rm skew}(\omega)$ denotes the skew-symmetric matrix associated with the vector $\omega\in\mathbb{R}^3$. 
	$\alpha_{x}$ denotes the velocity of $x\in{\cal X}$. If $\cal X$ is an Euclidean space then $\alpha_{x}=\dot{x}$. 
	$\quatern_{\star}^T = \begin{bmatrix}
		\bar{\quatern}_{\star} & \hat{\quatern}_{\star}
	\end{bmatrix}\in\mathbb{R}^4$ is the unit quaternion representation of $R_{\star}$ where  $\bar{\quatern}\in\mathbb{R}$ and $\hat{\quatern}\in\mathbb{R}^3$ with $\norm{\quatern_{\star}} =1$, and  ${\quatern_{\star}^{-1}}^T= \begin{bmatrix}
		\bar{\quatern}_{\star} & - \hat{\quatern}_{\star}
	\end{bmatrix}$. The vector  part of the  quaternion product $\quatern_\star \otimes \quatern_{\star\star}$ is denoted as $\hat{\quatern}_\star \ominus \hat{\quatern}_{\star\star}= \bar{\quatern}_{\star\star}\hat{\quatern}_{\star} + \bar{\quatern}_{\star}\hat{\quatern}_{\star\star} + \skewMat{\hat{\quatern}_{\star}}\hat{\quatern}_{\star\star}\inR^3$. If $\quatern_{\star\star}= \quatern_{\star}^{-1}$ then $\hat{\quatern}_\star \ominus \hat{\quatern}_{\star\star}=0$. $P_{\star}\in\mathbb{R}^3$ denotes the Cartesian coordinates of $\setR_{\star}$ origin. $\dot{P}_{\star},\ddot{P}_{\star}\in\mathbb{R}^3$ are is linear velocity and acceleration. The angular velocity and acceleration of $\setR_{\star}$ frame are $\omega_{\star},\dot{\omega}_{\star}\in\mathbb{R}^3$, respectively.

	\section{Handover Paradigm}
	\subsection{Handover Control Problem}
	In order to perform object handover efficiently, the robot reaching-motion toward the HOL shall be synchronized with the object's motion to achieve a \emph{one-shot continuous and smooth motion}. If the HOL is known in advance, the control formulation resumes to perform simply a set-point reaching task for the end-effector. Nevertheless, the HOL and its timing as well as the object orientation are highly versatile: they may change during the handover motion or between two successive processes depending on the human intention and her/his posture. Handing over an object can even be aborted on the human decision while moving. Hence, planning the robot motion is likely to fail in such conditions. Alternatively, a reactive control method is more suitable for its ability to adjust the robot motion in real-time. We denote by \emph{end-effector} the robot terminal link used to pick-up the object. It encompasses two-finger grippers, multi-finger hand and even a non-human like devices as a suction-cup or an electromagnetic gripper~\cite{pan2018haptics}.

	Our approach is to tackle the human-robot handover problem from a reactive closed-loop control perspective. We shall explain our approach when the human is the giver, then we show how it applies when s/he is the receiver. The HOL can be seen as an attractor toward which the object is converging (steered by the human). Since the HOL is not known a priori, the robot end-effector can track the object trajectory leading to a proactive motion such that when the object converges to the HOL, its trajectory becomes a point to which the end-effector converges. The same reasoning can be applied for the orientation. The remaining question is how to obtain the object trajectory in terms of pose $X_{\rm obj}\in\mathbb{R}^7$, linear and angular velocity $\alpha_{X_{\rm obj}}\in\mathbb{R}^6$ and linear and angular acceleration $\dot{\alpha}_{X_{\rm obj}}\in\mathbb{R}^6$? These terms are encompassed as the \emph{object full-state} 
	\begin{equation}\label{eq:object full-state}
		s_{\rm obj}=\begin{bmatrix}
			X_{\rm obj}\\ \alpha_{X_{\rm obj}} \\ \dot{\alpha}_{X_{\rm obj}}
		\end{bmatrix}\in\mathbb{R}^{19}.
	\end{equation} 
	Often, $s_{\rm obj}$ cannot be directly obtained by the sensors as generally they provide only $X_{\rm obj}$. Hence, an estimation of $s_{\rm obj}$ needs to be constructed from $X_{\rm obj}$ and it is denoted as 
	\begin{equation}\label{eq:observer full-state}
		s_{\rm obs}=\begin{bmatrix}
			X_{\rm obs}\\ \alpha_{X_{\rm obs}} \\ \dot{\alpha}_{X_{\rm obs}}
		\end{bmatrix}\in\mathbb{R}^{19}.
	\end{equation}
	To this end, the observation task is formulated as a PD controller that drives 
	$X_{\rm obs}$ toward $X_{\rm obj}$, e.g.,~\cite{paolillo2018ral}. While converging, the observation task outputs also $\alpha_{X_{\rm obs}}$ and $\dot{\alpha}_{X_{\rm obs}}$, e.g.~\cite{pham2018pami}. $s_{\rm obs}$ terms are then used as references for the trajectory-tracking task formulated for the end-effector. Note that the observation task allows to generate trajectories for the orientation tracking without the need for offline planning methods~\cite{sciavicco2000book}. The same approach can be adapted if the human is a receiver. In such case, the end-effector already holds the object and its full-state known by forward kinematics. Instead, the observation task constructs the full-state of the human hand~\cite{pham2018pami}. 
	
	Multi-objective task-space QP control has been extensively used as it enables to specify various tasks objectives while explicitly accounting for a set of convex constraints~\cite{bolotnikova2020roman}. In~\cite{bouyarmane2019tro}, one single QP controller can be formulated for multi-robots systems either decoupled or interacting with each other using contact forces. In particular, the pre-grasping handover phase can be suitably formulated using the multi-robots QP by considering the robotic arm and the object as two decoupled robots where the former is a redundant multi-body system where all the joints are actuated, and the latter as a floating-base rigid-body system. Moreover, the observation task is formulated for the object to compute $s_{\rm obs}$ and the trajectory-tracking task is formulated for the end-effector.   
	The following subsections explicit in details our approach.     
	\subsection{Background}
	Consider a $n$ degree-of-freedom redundant robotic arm such that its joint configuration is defined by $\actConf\in\mathbb{R}^n$. The robot is governed by the following equation of motion:
	\begin{equation}\label{eq:equation of motion}
		M(q)\ddot{q} + N(q,\dot{q}) = \tau + \actJac_{\rm c}^Tf,
	\end{equation}
	where $M(q)\in\mathbb{R}^{n\times n}$ is the inertia matrix, $N(q,\dot{q})\in\mathbb{R}^{n}$ encompasses Coriolis, centrifugal and gravity torques, $\tau\in\mathbb{R}^{n}$ is the joint-torque, $\actJac_{\rm c}\in\mathbb{R}^{6\times n}$ is the Jacobian at the contact point and $f\in\mathbb{R}^6$ is the contact wrench. 
	Let us consider the frame $\setR_{\rm ee}$ rigidly attached to the end-effector and which pose is   
	\begin{equation}\label{eq:end-effector pose}
		X_{\rm ee}=\begin{bmatrix}
			P_{\rm ee} \\ \quatern_{\rm ee}
		\end{bmatrix}\in \mathbb{R}^7.
	\end{equation}
	Its velocity and acceleration are given as 
	\begin{align}\label{eq:end-effector vel and acc}
		\alpha_{X_{\rm ee}}&=\begin{bmatrix}
			\dot{P}_{\rm ee} \\ \omega_{\rm ee}
		\end{bmatrix} = J_{\rm ee}\dot{q}\in \mathbb{R}^6, \\
		\dot{\alpha}_{X_{\rm ee}}&=\begin{bmatrix}
			\ddot{P}_{\rm ee} \\ \dot{\omega}_{\rm ee}
		\end{bmatrix} = J_{\rm ee}\ddot{q} + \dot{J}_{\rm ee}\dot{q}\in \mathbb{R}^6,
	\end{align} where $J_{\rm ee}\inR^{6\times n}$ is the end-effector Jacobian.

	Let us consider the object as a one-rigid-link robot with 6 DoF to which a frame $\frameObj$ is rigidly attached and which pose is 
	\begin{equation}\label{eq:object pose}
		X_{\rm obj}=\begin{bmatrix}
			P_{\rm obj} \\ \quatern_{\rm obj}
		\end{bmatrix}\in \mathbb{R}^7,
	\end{equation}

	where $X_{\rm obj}$ is assumed to be provided by a sensor. 
	Its velocity and acceleration are given as 
	\begin{align}\label{eq:obj vel and acc}
		\alpha_{X_{\rm obj}}&=\begin{bmatrix}
			\dot{P}_{\rm obj} \\ \omega_{\rm obj}
		\end{bmatrix}\in \mathbb{R}^6, \
		\dot{\alpha}_{X_{\rm obj}}=\begin{bmatrix}
			\ddot{P}_{\rm obj} \\ \dot{\omega}_{\rm obj}
		\end{bmatrix}\in \mathbb{R}^6.
	\end{align}
	The object structural and dimension properties are known.
	
	\subsection{Observation Task}\label{subsec:observation task}
	
	Let us consider an observed object to which a frame $\setR_{\rm obs}$ is rigidly attached which pose is 
	\begin{equation}
		X_{\rm obs} = 
		\begin{bmatrix}
			P_{\rm obs} \\ \quatern_{\rm obs}
		\end{bmatrix}\in \mathbb{R}^7 .
	\end{equation}
	
		Assuming that the object (linear and angular) velocity and acceleration~\eqref{eq:obj vel and acc} are not provided by the sensor (which is likely the case), the observation task  aims at constructing these non-measured states by estimating $s_{\rm obs}$ in~\eqref{eq:observer full-state}.
		 
		This is achieved by the observation task that steers $X_{\rm obs}$ toward $X_{\rm obj}$ by keeping the observation error $e_{\rm obs}$ as small as possible such that 
		\begin{align}\label{eq:observation error}
			e_{\rm obs} &= 
			\begin{bmatrix}
				P_{\rm obs}-P_{\rm obj} \\ \hat{\quatern}_{\rm obs} \ominus \hat{\quatern}_{\rm obj}^{-1}
			\end{bmatrix}\in \mathbb{R}^{6}.
		\end{align}
		The observation error velocity and acceleration are  given as
		\begin{align} 
			\alpha_{e_{\rm obs}} = \alpha_{X_{\rm obs}}, \ 
			\dot{\alpha}_{e_{\rm obs}} = \dot{\alpha}_{X_{\rm obs}}.
		\end{align}
		The observation task state is defined as 
		\begin{equation}
			\observTaskState = 	\begin{bmatrix}
				e_{\rm obs} \\	\alpha_{e_{\rm obs}}
			\end{bmatrix}  \in \mathbb{R}^{12}.
		\end{equation}
		Thus, the observation task is formulated as follows
		\begin{align}
			\label{eq:mu observartion task}\observTaskStateDot &= 
			\begin{bmatrix}
				0 & I \\ 0 &0 
			\end{bmatrix}\observTaskState + 
			\begin{bmatrix}
				0 \\ I
			\end{bmatrix} \mu_{\rm obs}, \ \mu_{\rm obs} =\dot{\alpha}_{X_{\rm obs}}
		\end{align}
		Choosing $\mu_{\rm obs}$ in~\eqref{eq:mu observartion task} a linear state feedback 
		\begin{equation}\label{eq:observation-task feedback law}
			\mu_{\rm obs} = - \left[K_{s_{\rm obs}} \ K_{d_{\rm obs}}\right]\observTaskState = -K_{\rm obs}\observTaskState,
		\end{equation}
		with $K_{s_{\rm obs}},K_{d_{\rm obs}}\in \mathbb{R}^{6\times6}$ are diagonal positive-definite matrices denoting the stiffness and damping gains; it yields to the observation-task closed-loop dynamics
		\begin{align}
			\label{eq:observation-task closed-loop}\observTaskStateDot &= A_{\rm obs}\observTaskState, \ A_{\rm obs} = 
			\begin{bmatrix}
				0 & I \\ -K_{s_{\rm obs}} & -K_{d_{\rm obs}}
			\end{bmatrix},
		\end{align}
		with $A_{\rm obs}$ Hurwitz~\cite{khalil2002NonLinearSystems}. Note that $\observTaskState$ only converges asymptotically to zero if the object is static ($X_{\rm obs}$ constant). However, choosing high gain values typically enables  a fast convergence and keeps $\norm{e_{\rm obs}}$ sufficiently small.
		
		The benefits of the observation task are three folds: (i) allows a bounded estimation of $s_{\rm obs}$ in~\eqref{eq:observer full-state} given~\eqref{eq:mu observartion task}--\eqref{eq:observation-task closed-loop}; (ii) $X_{\rm obj}$ is low-pass filtered by the closed-loop observation task dynamics~\eqref{eq:observation-task closed-loop}; (iii) enables an online generation of a smooth twice-differentiable trajectory\footnote{This is also known as a \emph{reference model}-based approach for trajectory references generation~\cite[Chapter 13]{khalil2002NonLinearSystems}. In addition, the trajectory feedforward terms are generated in real-time  since the observation task is updated at the same frequency of QP no matter the sampling-frequency of the sensor.} for the Cartesian and orientation coordinates of $\setR_{\rm obs}$ without the needs of an offline planning methods~\cite{sciavicco2000book}. The latter advantage is the \emph{core idea} presented in this paper: the observation task outputs trajectory references required by the trajectory-tracking task in order to achieve a proactive handover process. More importantly, theses two tasks are integrated in the multi-objective QP controller as shown in \cref{subsec:QP combination}. 		
		Once having $s_{\rm obs}$, we can compute the full-state for any other frame $\setR_{\star}$ attached to the object (for which the local pose $X_{\star}^{\rm obs}$ is known) by simply applying the classical kinematic relations for position, velocity and acceleration: 
		\begin{align}
			\label{eq:forward position}	P_{\star} &\!=\! P_{\rm obs} + R_{\rm obs}P_{\star}^{\rm obs}, \\
			\label{eq:forward velocity}	\dot{P}_{\star} &\! =\! \dot{P}_{\rm obs} + \skewMat{\omega_{\rm obs}} R_{\rm obs}P_{\star}^{\rm obs}, \\
			\begin{split}
				\label{eq:forward acceleration}	\ddot{P}_{\star} & \!=\! \ddot{P}_{\rm obs} \!+\!\skewMat{\dot{\omega}_{\rm obs}} R_{\rm obs}P_{\star}^{\rm obs} \!+\! \skewMat{\omega_{\rm obs}}^2 R_{\rm obs}P_{\star}^{\rm obs}
			\end{split},\\
			\label{eq:rotation matrix for grasp frame}R_{\star} &\!= \! R_{\rm obs}R_{\star}^{\rm obs}. 
		\end{align}
		Note that the rigid body assumption implies that $\dot{P}_{\star}^{\rm obs}=0$, $\dot{R}_{\star}^{\rm obs}=0$, $\omega_{\star} = \omega_{\rm obs}$ and $\dot{\omega}_{\star} = \dot{\omega}_{\rm obs}$.
		\begin{figure}
			\centering
			\includegraphics[width=0.7\columnwidth]{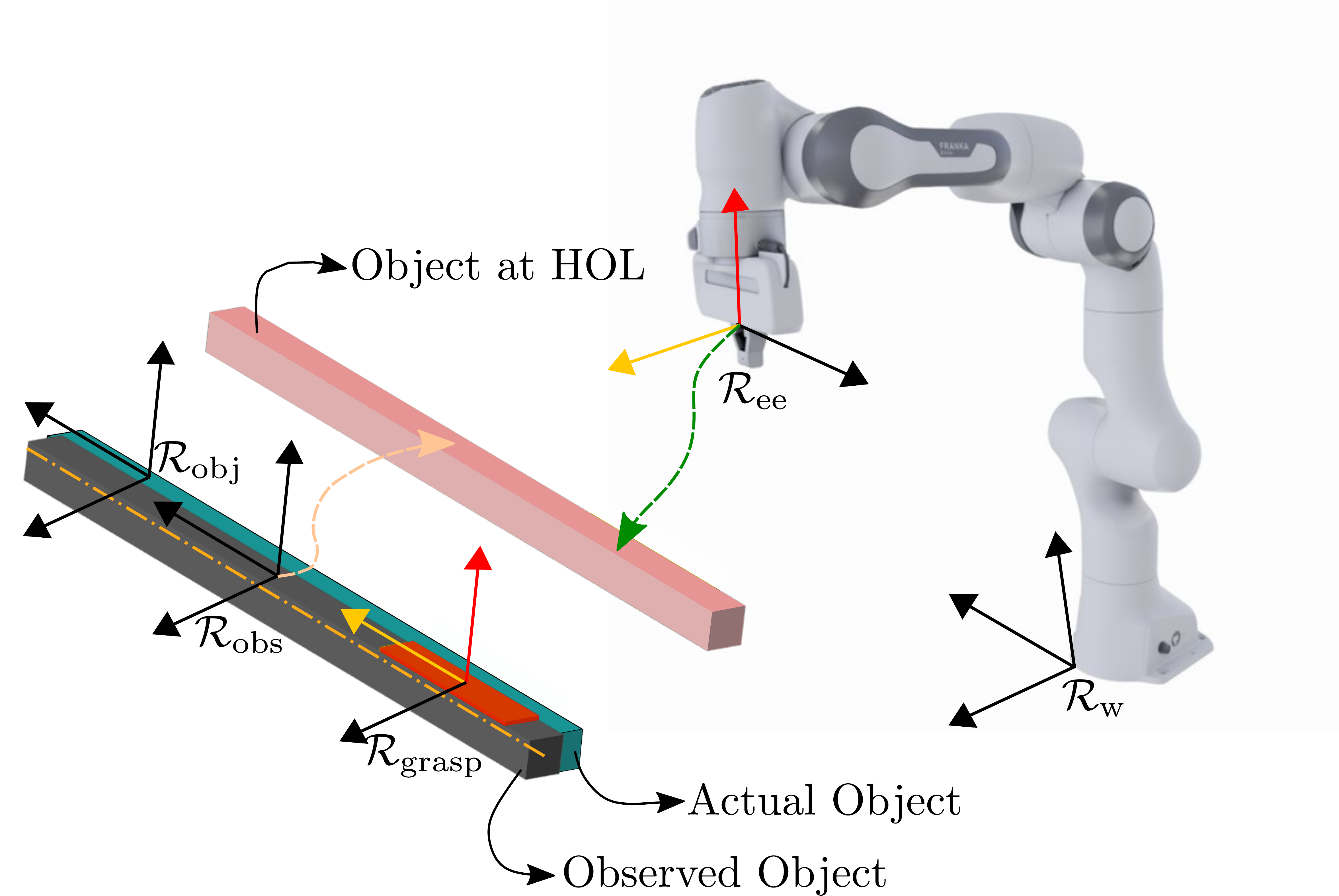}
			\caption{Illustrative scheme showing the  frames $\frameWorld$, $\frameEndEff$, $\frameObj$, $\frameObs$ and $\frameGrasp$. The observed object is tracking the actual object giving the data from sensor thanks to the observation task. The end-effector tracks the observation task state yielding an anticipatory motion toward the HOL where it ultimately meets the object. The colored unit vectors in $\frameEndEff$ track their corresponding in $\frameGrasp$. For instance, if the object is cylindrical, only the yellow unit vectors must be aligned.} 
		\label{fig:robot and object}
	\end{figure}
	\subsection{Trajectory-Tracking Task}

	Let $\setR_{\rm grasp}$ be the grasping frame 
	rigidly attached to the object and which pose is defined as  
	\begin{equation}
		X_{\rm grasp} = \begin{bmatrix}
			P_{\rm grasp} \\ \quatern_{\rm obs}
		\end{bmatrix}\in\mathbb{R}^7,
	\end{equation}
	such that  $P_{\rm grasp}\in\mathbb{R}^3$ denotes the coordinates of the point where the end-effector grasps the object and  is computed from \eqref{eq:forward position}, and $\quatern_{\rm obs}$ is obtained from $R_{\rm grasp}$ in~\eqref{eq:rotation matrix for grasp frame}. $\setR_{\rm grasp}$ velocity and acceleration are computed as 
	\begin{align}\label{eq:grasp vel and acc}
		\alpha_{X_{\rm grasp}}&=\begin{bmatrix}
			\dot{P}_{\rm grasp} \\ \omega_{\rm obs}
		\end{bmatrix}\in \mathbb{R}^6, \
		\dot{\alpha}_{X_{\rm grasp}}=\begin{bmatrix}
			\ddot{P}_{\rm grasp} \\ \dot{\omega}_{\rm obs}
		\end{bmatrix}\in \mathbb{R}^6.
	\end{align}
	Let us define the trajectory-tracking task error as 
	\begin{align}\label{eq:grajectory tracking error}
		e_{\rm tt} &= 
		\begin{bmatrix}
			P_{\rm ee}-P_{\rm grasp} \\ \hat{\quatern}_{\rm ee} \ominus \hat{\quatern}_{\rm grasp}^{-1}
		\end{bmatrix}\in \mathbb{R}^{6},
	\end{align}
	and which derivative is given as 
	\begin{align}\label{eq:grajectory tracking errorDot}
		\alpha_{e_{\rm tt}} &= 
		\begin{bmatrix}
			\dot{P}_{\rm ee}-\dot{P}_{\rm grasp} \\ \omega_{\rm ee} - \omega_{\rm obs}
		\end{bmatrix}= \alpha_{X_{\rm ee}}-\alpha_{X_{\rm grasp}}\in \mathbb{R}^{6}.
	\end{align}
	Let us denote the trajectory-tracking task state 
	\begin{align}\label{eq:grasping-point-tracking task state}
		\trajTrackTaskState = 
		\begin{bmatrix}
			e_{\rm tt} \\ \alpha_{e_{\rm tt}}
		\end{bmatrix}\in\mathbb{R}^{12} .
	\end{align}
	Hence, the trajectory-tracking task dynamics is formulated 
	\begin{align}
		\trajTrackTaskStateDot &= 
		\begin{bmatrix}
			0 & I \\ 0 &0 
		\end{bmatrix}\trajTrackTaskState+ 
		\begin{bmatrix}
			0 \\ I
		\end{bmatrix} \mu_{\rm tt} ,\\
		\label{eq:mu grasping-point-tracking task}\mu_{\rm tt} &=\dot{\alpha}_{X_{\rm ee}}-\dot{\alpha}_{X_{\rm grasp}} = J_{\rm ee}\ddot{q} +  \dot{J}_{\rm ee}\dot{q} - \dot{\alpha}_{X_{\rm grasp}}.
	\end{align}
	Then, choosing $\mu_{\rm tt}$ in~\eqref{eq:mu grasping-point-tracking task} as 
	\begin{equation}\label{eq:grasping-point-tracking task feedback law}
		\mu_{\rm tt} = - \left[K_{s_{\rm tt}} \ K_{d_{\rm tt}}\right]\trajTrackTaskState = -K_{\rm tt}\trajTrackTaskState,
	\end{equation}
	with $K_{s_{\rm tt}},K_{d_{\rm tt}}\in\mathbb{R}^{6\times6}$ are diagonal positive-definite matrices denoting the stiffness and damping gains; it yields to the following trajectory-tracking task closed-loop dynamics
	\begin{equation}
		\label{eq:grasping-point-tracking-task closed-loop}\trajTrackTaskStateDot
		= A_{\rm tt}\trajTrackTaskState, \ A_{\rm tt} = 
		\begin{bmatrix}
			0 & I \\ -K_{s_{\rm tt}} & -K_{d_{\rm tt}}
		\end{bmatrix}.
	\end{equation}
	where $A_{\rm tt}$ is Hurwitz. This enables a global asymptotic convergence of  $\trajTrackTaskState$ to the origin~\cite{khalil2002NonLinearSystems}. Commonly, the trajectory is planned such that the initial reference trajectory pose is as close as possible to the current end-effector pose $X_{\rm ee}(t_0)$ which indeed ensures that the tracking error $e_{\rm tt}(t_0)$ is small and thereby enforces the end-effector pose to stick on trajectory forward in time. However, when the trajectory starts far from the initial end-effector pose, the latter converges to the trajectory  with an asymptotic decay of the tracking error $\trajTrackTaskState$. This property enables the anticipatory motion of the end-effector which moves proactively toward the object grasping position (see \cref{fig:robot and object}).
	\subsection{Posture Task}
	The posture task is mainly intended to solve the remaining redundant states $(q,\dot{q})$ and keep them bounded.	
	Let  $q_{\rm ref}\in\mathbb{R}^n$ be a given reference posture designed generally to represent a suitable robot posture (e.g., elbow up). Let us define the posture task state as 
	\begin{equation}
		\eta_{\rm pos} = \begin{bmatrix}
			e_{\rm pos} \\ \dot{e}_{\rm pos}
		\end{bmatrix} = 
		\begin{bmatrix}
			q - q_{\rm ref} \\ \dot{q}
		\end{bmatrix}.
	\end{equation} 
	The posture task dynamics is thereby 
	\begin{align}
		\label{eq:mu posture task}\dot{\eta}_{\rm pos} &= \begin{bmatrix}
			0& I \\ 0&0
		\end{bmatrix}\eta_{\rm pos} + \begin{bmatrix}
			0 \\ I 
		\end{bmatrix}\mu_{\rm pos}, \ \mu_{\rm pos}=\ddot{q} .		
	\end{align}
	Choosing $\mu_{\rm pos}$ in~\eqref{eq:mu posture task} as 
	\begin{equation}
		\mu_{\rm pos} = -\begin{bmatrix}
			K_{s_{\rm pos}} & K_{d_{\rm pos}}
		\end{bmatrix}\eta_{\rm pos} = -K_{\rm pos},
	\end{equation}
	leads to the posture task closed-loop dynamics
	\begin{equation}\label{eq:posture task closed-loop dynamics}
		\dot{\eta}_{\rm pos} = A_{\rm pos}{\eta}_{\rm pos}, \ A_{\rm pos} = \begin{bmatrix}
			0& I \\ -K_{s_{\rm pos}} & -K_{d_{\rm pos}}
		\end{bmatrix},
	\end{equation}
	where $A_{\rm pos}$ is Hurwitz given that the stiffness and damping gains $K_{s_{\rm pos}}$, $K_{d_{\rm pos}}$ are diagonal positive-definite. \cref{eq:posture task closed-loop dynamics} yields asymptotic convergence of $\eta_{\rm pos}$ to the origin. However, this is only the case when the posture task is not in conflict with the other tasks and constraints among which the posture task has the lowest priority. Alternatively, $e_{\rm pos}$ is only ensured to be uniformly
	ultimately bounded~\cite{bouyarmane2018tac}.
	\subsection{Multi-Tasks QP Formulation}\label{subsec:QP combination}
	Let us denote $\chi^T=\begin{bmatrix}
		\ddot{q}^T & \dot{\alpha}_{\rm obs}^T & f^T
	\end{bmatrix}\in\mathbb{R}^{n+6+3}$ the vector encompassing the linear and angular acceleration of the observed rigid-body object, the joint-acceleration of the robotic arm, and the contact force.
	Given the affinity of~\eqref{eq:mu observartion task} w.r.t $\dot{\alpha}_{X_{\rm obs}}$, that of~\eqref{eq:mu grasping-point-tracking task}\eqref{eq:mu posture task} w.r.t $\ddot{q}$ and that of~\eqref{eq:equation of motion} w.r.t $\ddot{q}$ and $f$, then we can combine all the tasks and constraint in a single weighted-prioritized QP formulation 
	\begin{subequations}\label{eq:QP formulation}
		\begin{align}
			\label{eq:QP cost-function}\underset{\chi}{\min} \sum_{i}&w_i\norm{E_{\chi}^i\chi+ F_\chi^i}^2 \\
			\label{eq:QP constraint}\text{S.t:~} &C_{\chi}\chi\leq d_{\chi}
		\end{align}
	\end{subequations}
	where $w_i>0$ is the associated weight to each task $i$.
	QP formulation~\eqref{eq:QP formulation} enables to: (i) combine the different competing tasks in~\eqref{eq:QP cost-function} by settling a soft prioritization scheme; (ii) account for unilateral constraints~\eqref{eq:QP constraint} that  embed limitations like joint-position, velocity, acceleration and torque limits, as well as safety aspects like (self)-collision avoidance to ensure safe grasping~\cite{djeha2020ral}; (iii) find optimal $\ddot{q}$ that generates the robotic arm motion that achieve \emph{at best} all the tasks while fulfilling all the constraints; and (iv)  handle multi-robot control where the robotic arm and the object are considered as two distinct robots entities which opens the possibility to add other robots to achieve the handover (e.g., bi-arm handover) while still using one compact formulation~\cite{bouyarmane2019tro}. 
	\begin{figure}
		\centering
		\includegraphics[width=0.3\columnwidth]{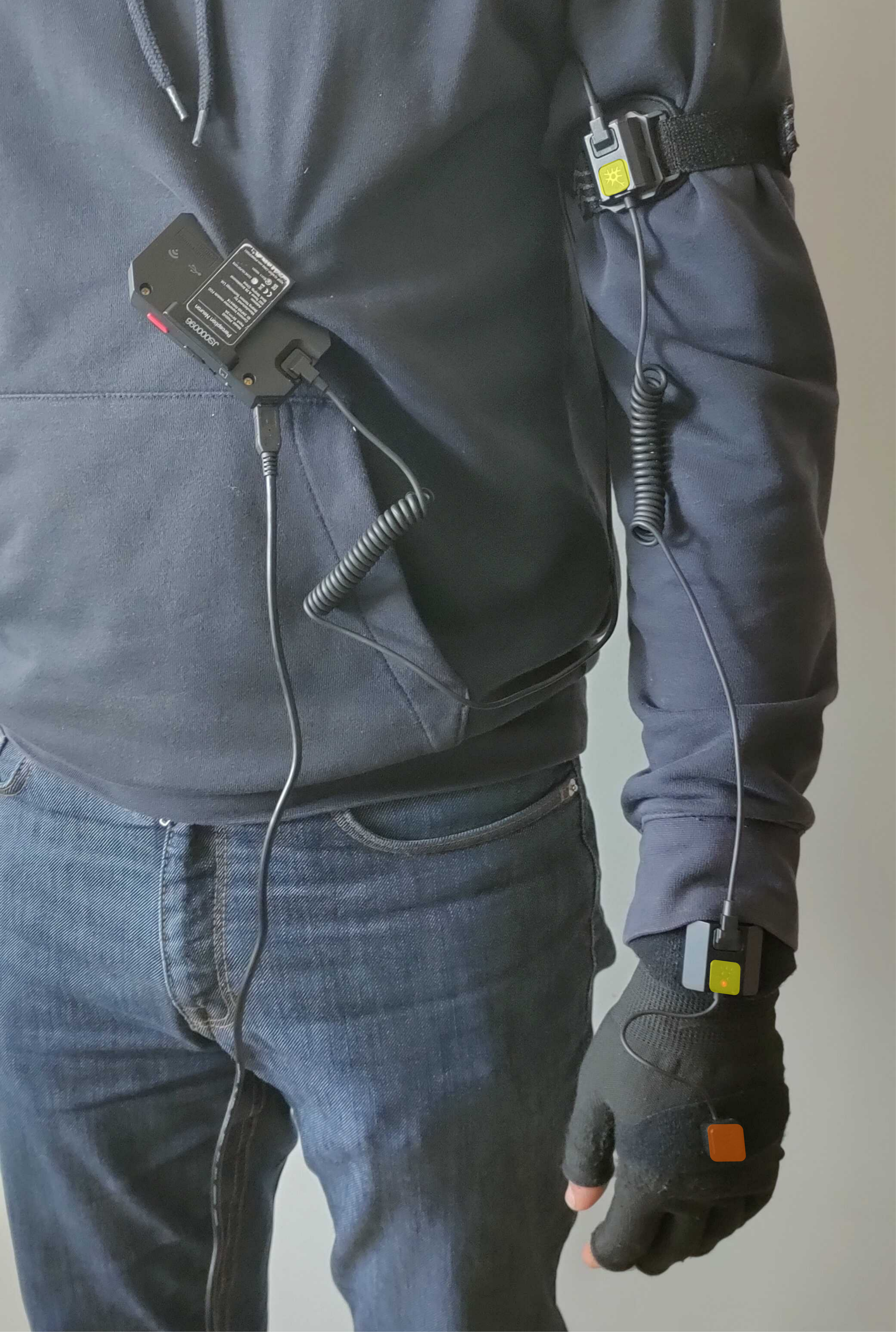}
		\caption{Perception Neuron motion capture sensor-suit used for the handover experiments and mounted on the left arm. Three IMUs (highlighted in colored squares) mounted on the left hand are used to provide an estimation of the hand pose (the orange square). }
		\label{fig:mocap_suit}
	\end{figure}
	\section{Experimental Result}
\begin{figure}
		\centering
		\includegraphics[width=0.91\columnwidth]{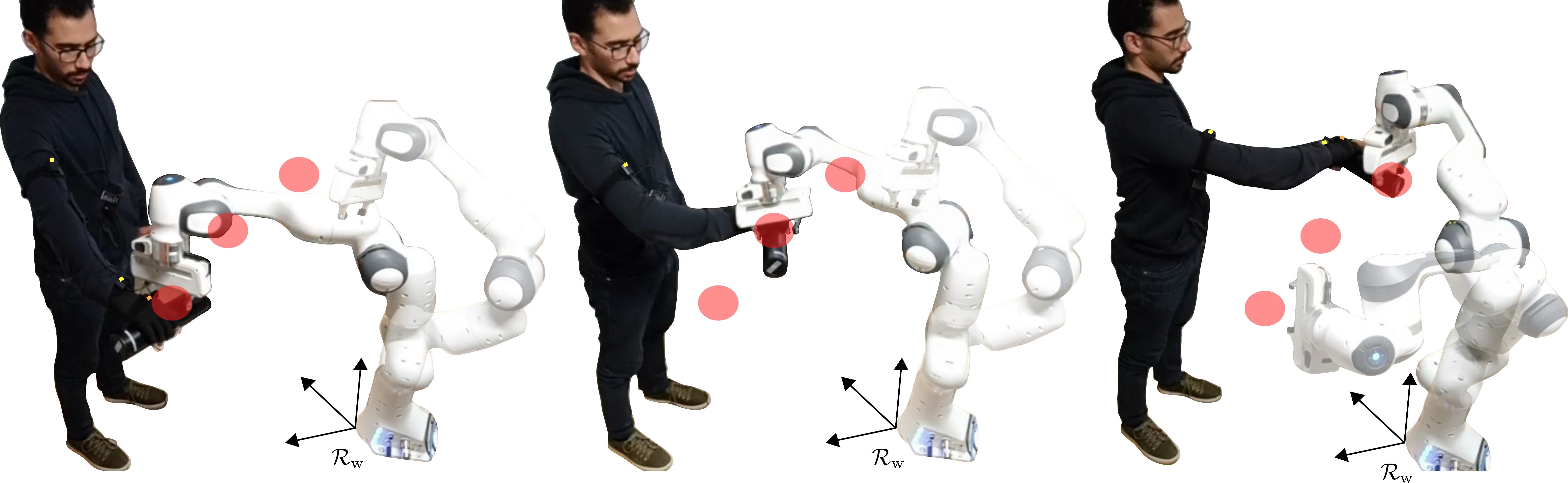}
\caption{Different HOLs (red) desired by the human operator. The object is a black cylindrical container. The transparent robots represent the initial configuration.}
\label{fig:HOL}
\end{figure}

\begin{figure}
\centering
\includegraphics[width=0.4935\columnwidth]{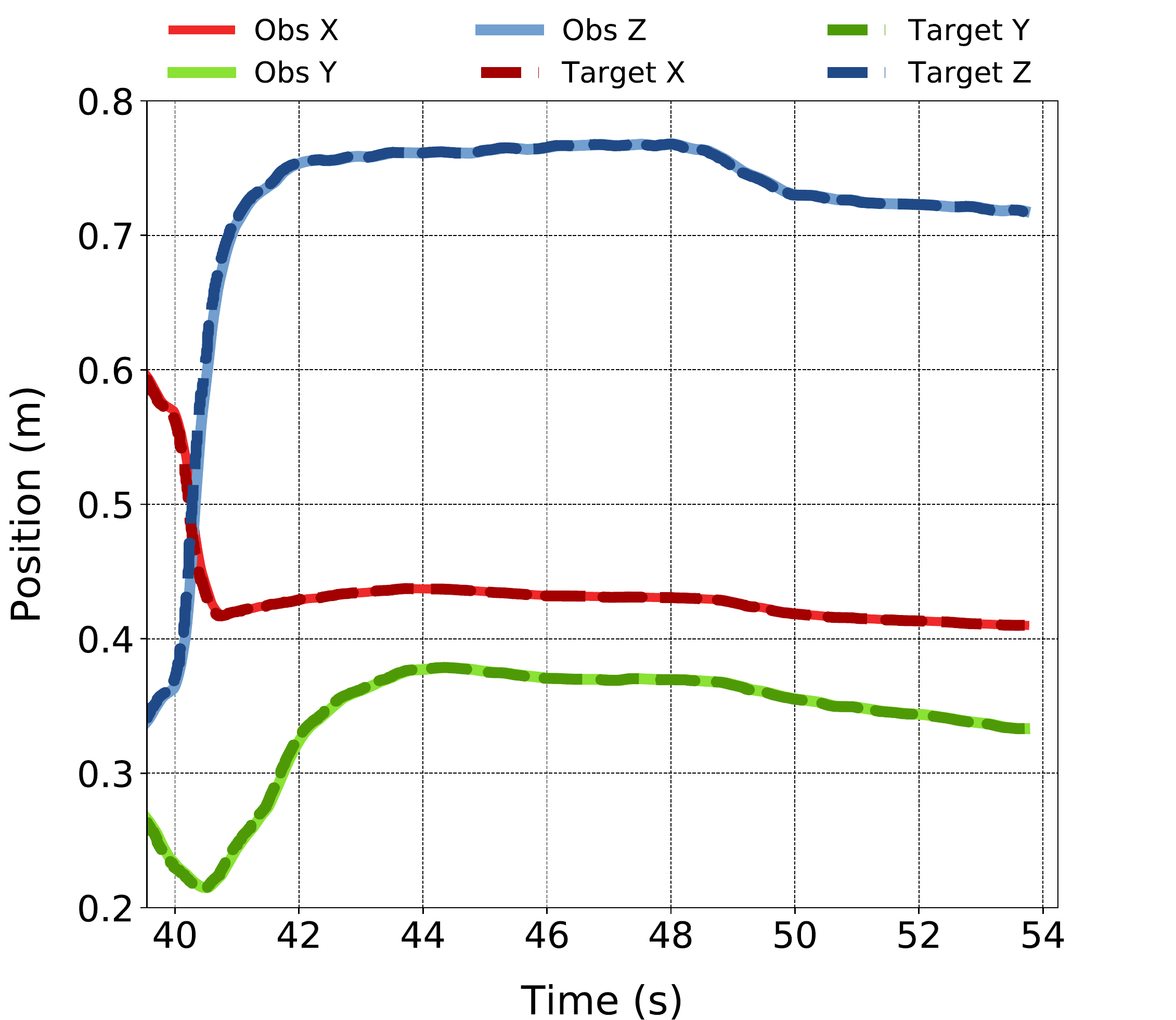}
\includegraphics[width=0.4935\columnwidth]{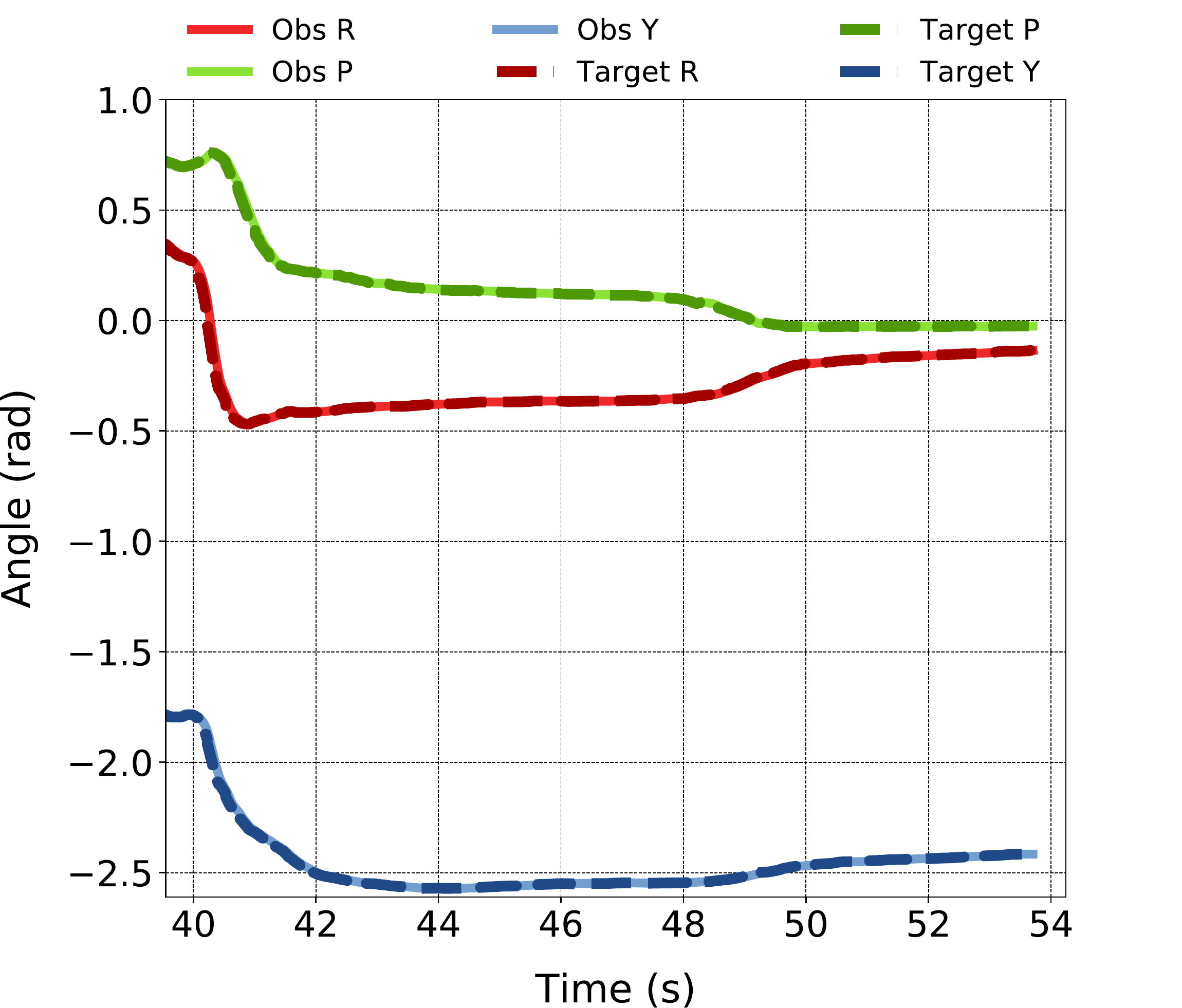}
\caption{Position (left) and orientation (right with RPY angles) of $\frameObj$ (dashed) and its estimation $\frameObs$ (solid) obtained by the observation task.}
\label{fig:obs}
\end{figure}

\begin{figure}
\centering
\includegraphics[width=0.4935\columnwidth]{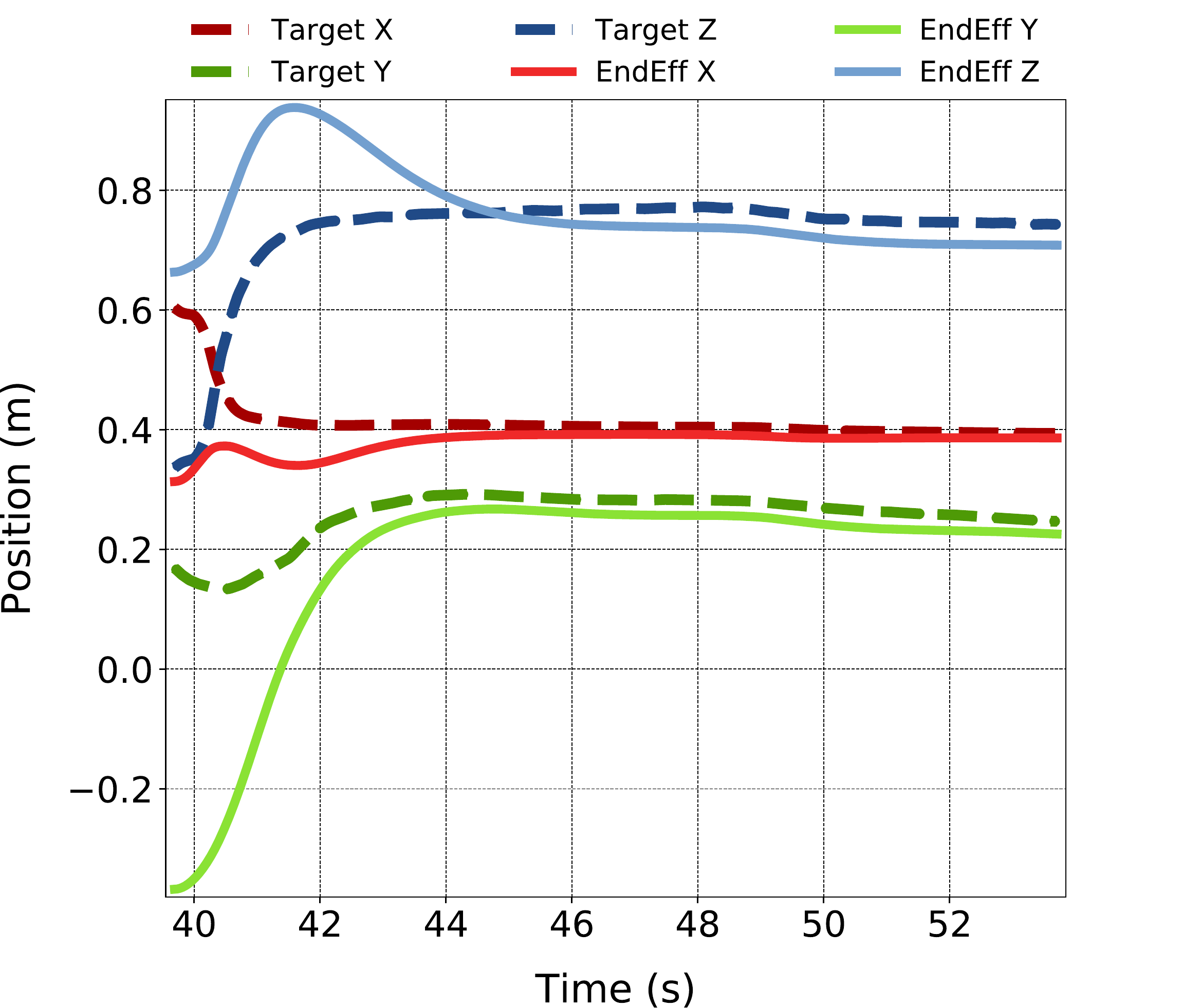}
\includegraphics[width=0.4935\columnwidth]{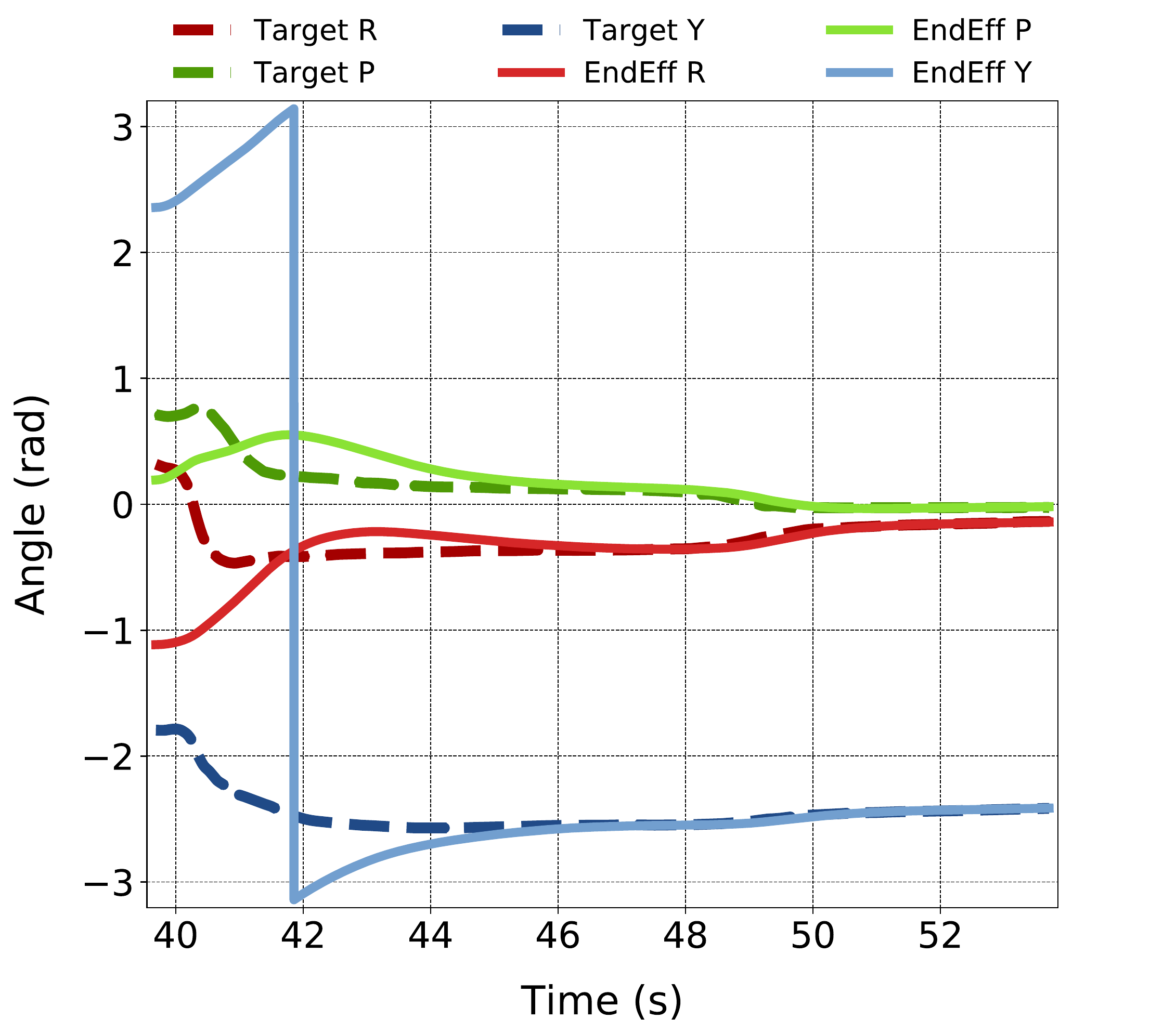}
\caption{Position (left) and orientation (right with RPY angles) of the frames $\frameGrasp$ (dashed) and  $\frameEndEff$ (solid)  obtained by the trajectory-tracking task. The sudden variation in the yaw angle is sound because the angle is bounded between $-\pi$ and $\pi$. This does not induce any singularity issue since the orientation is implemented using unit quaternions. }
\label{fig:trajTrack}
\end{figure}

\begin{figure}
\centering
\includegraphics[width=0.65\columnwidth]{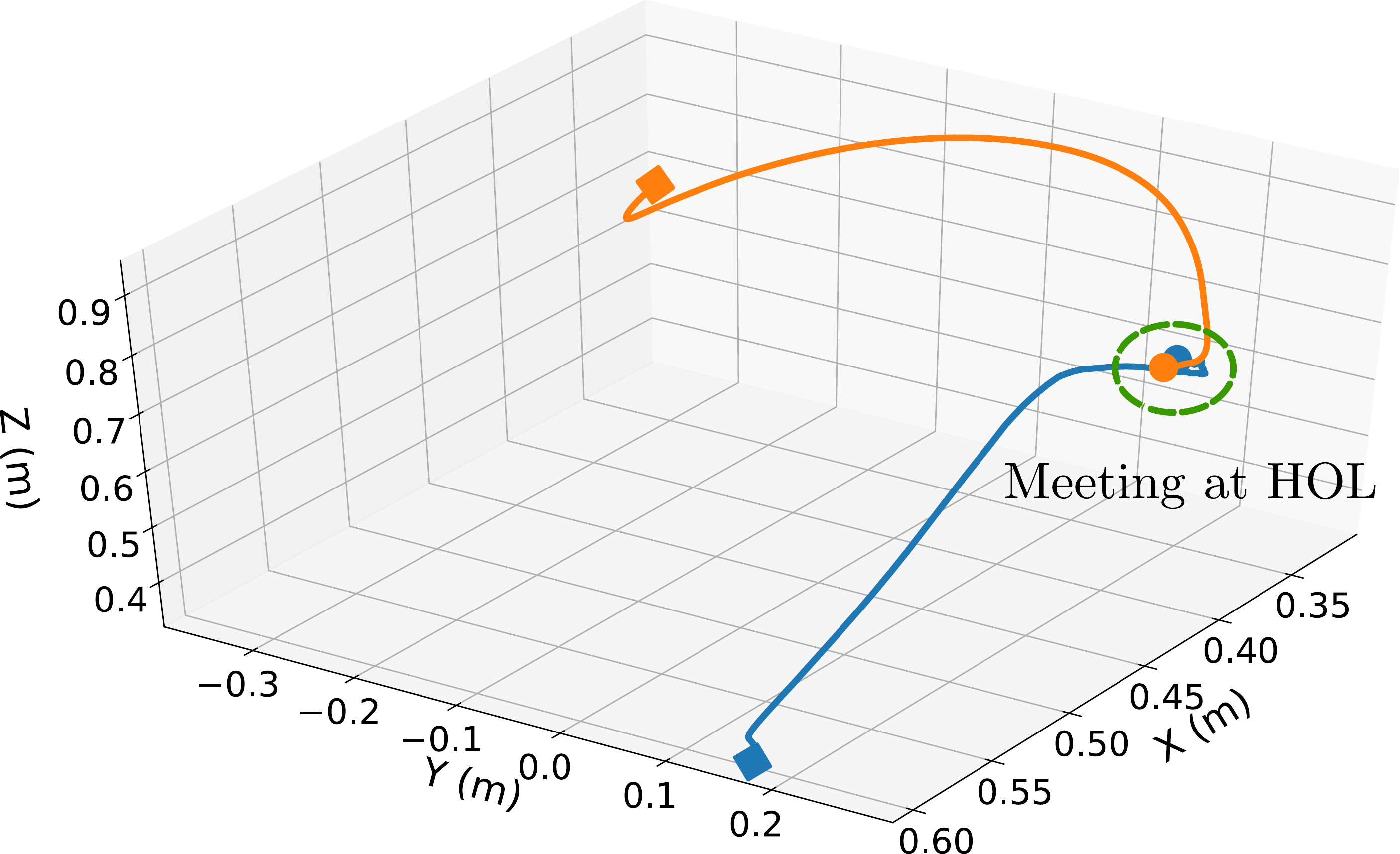}
\caption{3D trajectories of $\frameGrasp$ (blue) and $\frameEndEff$ (orange). The initial and terminal points are denoted with a square and a circle, respectively.  }
\label{fig:3D plot}
\end{figure}

	For demonstration, we implemented our approach on the open-source code implementation of the QP controller interface \mcrtc\footnote{\url{https://jrl-umi3218.github.io/mc_rtc/index.html}}. Based on the embedded sensors data, \mcrtc~builds and solves the QP problem at each control cycle ($1$~ms). Experiments are conducted using  7-DoF robotic arm Panda. Perception Neuron\footnote{\url{https://neuronmocap.com/}} sensor-suit has been used for the pose measurement (\cref{fig:mocap_suit}). An API provides the hand pose in an arbitrarily fixed-frame at a frequency of $60$~Hz.  A calibration step is necessary to correctly coincides the sensor frame with $\frameWorld$. 
	The calibration process is repeated at the beginning of each handover experiment to avoid drifting. When the human is a giver, the object is assumed to be rigidly attached to the hand, and thereby its pose can be computed based on the hand's one. 
	Several handover scenarios have been performed where the robot starts from different configurations. In addition, multiple HOLs have been tried (\cref{fig:HOL}). The observation task gains have been set to high values $K_{s_{\rm obs}} = 1500I$,  $ K_{d_{\rm obs}} = 2\sqrt{K_{s_{\rm obs}}}$ to ensure an accurate object pose estimation $s_{\rm obj}$ \eqref{eq:object full-state}. \cref{fig:obs} shows how $X_{\rm obs}$ tracks $X_{\rm obj}$. The high observation task gains allow an accurate pose tracking. Concurrently, it allows obtaining a good estimation of $\alpha_{X_{\rm obj}}$ and $\dot{\alpha}_{X_{\rm obj}}$ by computing $\alpha_{X_{\rm obs}}$ and $\dot{\alpha}_{X_{\rm obs}}$ from \eqref{eq:observation-task closed-loop}.  $s_{\rm obj}$ is then forwarded to the trajectory-tracking task where the gains of the orientation part have been chosen twice bigger than those of the translation to ensure that  end-effector reaches the object while its orientation has already converged to the target. The stiffness of the translation part is fixed at  $I$ and the corresponding damping  $2\sqrt{I}$. From \cref{fig:trajTrack}, the object and the end-effector meet seamlessly and smoothly at the HOL without being specified \emph{a priori}, see \cref{fig:3D plot}. Furthermore, the end-effector adapts reactively to the variation of the grasping point as shown in the attached video.
\section{Conclusion}
We propose an original new approach for human-robot handovers formulated using task-space optimization controller. The core idea is based on a novel implementation of tasks interdependency, which consists in providing the state of a task (of estimation nature) as an input to another task so that both meet at the HOL without explicit time or object configuration specification. Our experimental results confirmed very promising performances of simple handovers focusing mainly on the reaching phase.
This new approach raised very promising novel features of the task-space optimization control schemes. Extension in terms of functionalities and theoretical investigation on how observer-tasks can be embedded in a unified task-space observer/control framework through task interdependencies and constraints between task errors opens perspectives in embedding scheduling in the task-space control formalism. Our on-going and future research is focusing on this issue together with complete-phases and more complex handovers considering force cues.

\bibliographystyle{ieeetr}
\bibliography{biblio.bib}	
\end{document}